# BGF-YOLO: Enhanced YOLOv8 with Multiscale Attentional Feature Fusion for Brain Tumor Detection

Ming Kang, Chee-Ming Ting$^{(\boxtimes)}$, Fung Fung Ting, and Raphaël C.-W. Phan

School of Information Technology, Monash University, Malaysia Campus,
Subang Jaya, Malaysia
ting.cheeming@monash.edu

**Abstract.** You Only Look Once (YOLO)-based object detectors have shown remarkable accuracy for automated brain tumor detection. In this paper, we develop a novel BGF-YOLO architecture by incorporating Bi-level routing attention, Generalized feature pyramid networks, and Fourth detecting head into YOLOv8. BGF-YOLO contains an attention mechanism to focus more on important features, and feature pyramid networks to enrich feature representation by merging high-level semantic features with spatial details. Furthermore, we investigate the effect of different attention mechanisms and feature fusions, detection head architectures on brain tumor detection accuracy. Experimental results show that BGF-YOLO gives a 4.7% absolute increase of mAP$_{50}$ compared to YOLOv8x, and achieves state-of-the-art on the brain tumor detection dataset Br35H. The code is available at https://github.com/mkang315/BGF-YOLO.

## 1 Introduction

Detecting brain tumors in their early stages can lead to more effective treatments and better prognoses. Therefore, brain tumor detection is a critical aspect of medical diagnostics. Magnetic Resonance Imaging (MRI) is the best imaging test to visualize the brain and detect tumors. The You Only Look Once (YOLO) series has been demonstrated to detect brain tumors accurately. Kang et al. [9] proposed RCS-YOLO–a novel YOLO architecture with reparameterized convolution based on channel shuffle–on brain tumor detection and achieved a balance between accuracy and speed.

The YOLOv8 architecture [8] is mainly composed of the backbone and head parts, in which the neck is included in the head part. The backbone part, which is used for feature extraction, contains Conv, C2f (shortcut), and Spatial Pyramid Pooling Fast (SPPF) modules. The Conv, that is ConvBiSiLU (or CBS), and SPPF are the same as those in the YOLOv5 [7] architecture [27], where Conv is used to perform convolution operation on the input images and assist C2f (shortcut) in feature extraction and SPPF enables adaptive-sized output. The C2f (shortcut) module is a lightweight convolutional structure compared to the C3 module in YOLOv5. Thus, the gradient flow of the model is enriched by connecting more branches across layers. Therefore, the more vital feature representation ability is enabled. The C2f (shortcut) module enhances the ability to express features through dense and residual structures, which changes the number of channels through split and concatenate operations according to scaling coefficients to reduce computational complexity and model capacity. The SPPF module at the end of the backbone part increases the sensitivity and captures the feature information of different levels in the images. The structures of Feature Pyramid Networks (FPN) [15] and Path Aggregation Network (PANet) [16] are used for multiscale feature fusion in the neck part. The FPN-PANet structure and C2f (without shortcut) modules fuse feature maps of different scales from the three stages of the backbone, aggregating shallow information to deep features. The head part employs a decoupled-head structure with a classification and regression (i.e., localization) prediction end to alleviate the conflict between classification and regression tasks, and an anchor-free mechanism to improve the detection of objects with irregular height and width. For bounding box classification, YOLOv8 employs binary cross-entropy loss while varifocal loss [32] is an alternative option. It can better handle the category imbalance situation and improve detection accuracy. For bounding box regression, YOLOv8 employs distribution focal loss [11–13] to overcome the problem of category imbalance and background category, allowing the network to quickly focus on the distribution of locations close to the object. It also uses the Complete Intersection over Union (CIoU) loss function [36] to alleviate the overlap between predicted and ground truth boxes. Nevertheless, some improvements potentially boost detection performance in the context of object detection in specific domains, which can still pose challenges using vanilla YOLOv8.

Recent improvements of YOLOv8 focus on attention mechanisms, multiscale feature fusion networks, and regression loss. Multi-Head Self-Attention mechanism was employed in MHSA-YOLOv8 [14]. A lightweight YOLOv8 [30] was proposed by combining a dual-path gated attention and feature enhancement module with the original YOLOv8s. An improved YOLOv8 with the neck structure of Asymptotic Feature Pyramid Network (AFPN) [31] was proposed in [5]. UAV-YOLOv8 [23] utilized the BiFormer block [37], focal fasternet blocks, and Wise-IoU (WIoU) [20] within the YOLOv8. Another improved YOLOv8 [35] also added Biformer in the backbone of YOLOv8 for insulator fault detection. DCA-YOLOv8 [29] employed deformable convolution and Coordinate Attention (CA) [3] within YOLOv8 for fast cattle detection. CSS-YOLO [17] respectively introduced the Swin Transformer and Convolution Block Attention Module (CBAM) [26] into YOLOv8's backbone and neck.

In this paper, we propose a novel model called BGF-YOLO, which enhances the detection performance of YOLOv8 by incorporating Bi-level Routing Attention (BRA) [37], Generalized-FPN (GFPN) [6], and Fourth detecting head. The contributions of this work are summarized as follows: 1) We reconstruct the original neck part of YOLOv8 with a structured feature fusion network based on GFPN to facilitate effective feature fusion at different levels. 2) We leverage BRA for both dynamic and sparse attention mechanisms to focus on more salient features and reduce feature redundancy. 3) We add a fourth detecting head to enrich the scales of anchor boxes and optimize regression loss for detection. 4) To our best knowledge, this is the first use of enhanced YOLOv8 for brain tumor detection. The proposed modifications significantly improve tumor detection compared to YOLOv8. We also assess the effects of using various attention mechanisms, feature pyramid networks, and regression losses on the detection performance.

## 2  Methods

Figure 1 illustrates the architecture of the proposed BGF-YOLO. BGF-YOLO is characterized by a very deep and heavy neck, in contrast to the lightweight neck & head part in the YOLOv8. The backbone and head parts of the proposed network are based on those of YOLOv8. Details of each part of the BGF-YOLO network structure are described in this section.

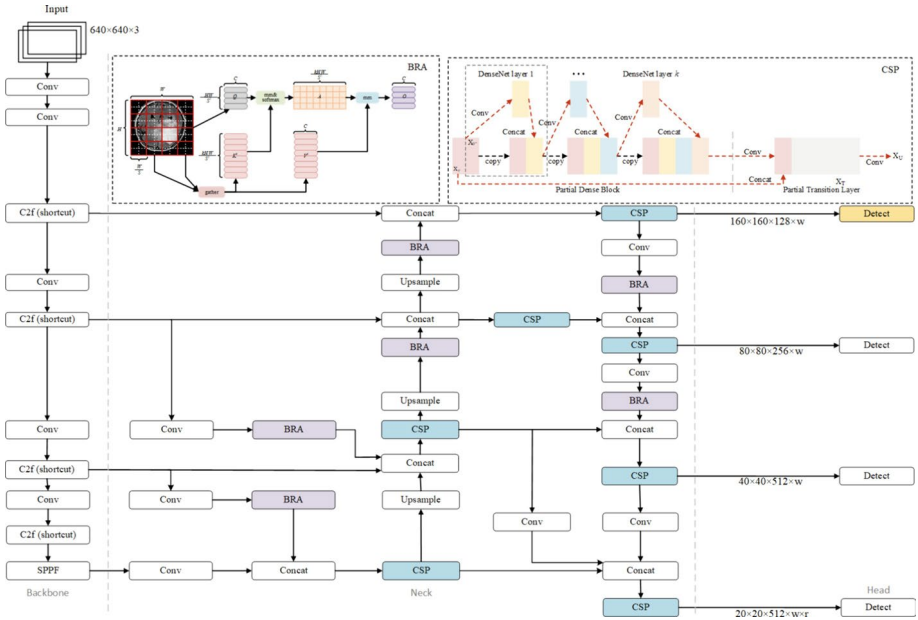

**Fig. 1.** Overview of BGF-YOLO. The architecture of BGF-YOLO is based on YOLOv8 and incorporates new modules colored Bi-level Routing Attention (BRA) [37], Cross Stage Partial DenseNet (CSP) [21], and enhanced detecting head. Conv, C2f (shortcut), SPPF, Concat, Upsample, and Detect are existing modules in the original YOLOv8 architecture [10].

### 2.1  Enhanced GFPN for Multilevel Feature Fusion

FPN was first introduced to address the hierarchical feature fusion issue of convolutional neural networks (CNNs) and has been proven to effectively enhance the capabilities of deep learning models in handling object detection tasks, especially for detecting objects at various scales. PANet has been used to strengthen feature propagation and encourage information reuse, thus improving the feature pyramid's representational power. Bidirectional FPN (BiFPN) [19] adds bottom-up pathways to FPN with only a top-down pathway, resulting in bidirectional cross-scale connections to harness multiscale features efficiently. Generalized-FPN (GFPN) [6] employs the structure of dense link and queen fusion to produce better-fused features and uses the concat operation instead of the sum to perform feature fusion to reduce loss of information. AFPN uses an adaptive spatial fusion in an asymptotic process, which initially fuses two low-level features, then higher-level features, and finally top-level features, to enhance the significance of pivotal levels and mitigate the impact of contradictory information from different objects.

FPN and PANet were later used for multiscale feature fusion in the necks of YOLOv5 and YOLOv8. The difference in the neck parts between YOLOv5 and YOLOv8 is that the C2f (without shortcut) modules in YOLOv8 replace the C3 modules in YOLOv5 at the upsampling stage. FPN first extracts the feature maps in the CNNs and then uses upper sampling and coarse-grained feature maps to achieve the fusion of the feature maps in a top-down manner. In contrast, PANet fuses feature maps from bottom to top to ensure spatial information is accurately preserved. However, the combination of FPN and PANet can only support top-down and bottom-up feature fusion. The structures of BiFPN, AFPN, and GFPN are supposed to quickly integrate features into various levels and improve the effect of feature fusion by adding more levels to meet the needs of feature fusion at different levels.

We modify the structure of the FPN-PANet in the YOLOv8 to achieve multilevel feature fusion among different layers by strengthening the multipath fusion of the networks. Inspired by the GFPN and reparameterized GFPN-based DAMO-YOLO [28], we utilize Cross Stage Partial DenseNet (CSP) [21] to add skip connections and simultaneously share dense information across various spatial scales and non-adjacent levels of latent semantics by replacing C2f (without shortcut) and combining with Conv. This allows the model to handle both high-level semantic information and low-level spatial information with equal importance in the neck part. Brain tumor images can exhibit variability in terms of tumor size, shape, and location. Cross-stage connections allow the network to better adapt to this variability by incorporating features from different scales, improving the model's robustness to diverse tumor characteristics.

## 2.2  BRA-Based Attentional Feature Fusion

The main idea of multiscale feature fusion networks in the neck part is to fuse feature maps extracted from different network layers to improve object detection performance at multiple scales. However, the feature fusion layer in YOLOv8 still suffers from the problem of redundant information from different feature maps. To overcome this limitation, we consider incorporating an attention mechanism to the feature fusion process in the YOLOv8 model.

The attention mechanism was originally proposed to weigh the importance of specific features relative to others. In the context of computer vision, there are five attention mechanisms that have great potential in improving the performance of object detection: Squeeze-and-Excitation (SE) [4], CBAM, Efficient Channel Attention (ECA) [22], CA, Receptive-Field Attention (RFA) [34], and BRA. The differences among them are that the SE and ECA belong to channel attention, the RFA and BRA handle spatial attention, and the CBAM and CA facilitate channel and spatial attention. SE is to adaptively recalibrate channel-wise feature responses by explicitly modeling the interdependencies between the channels of convolutional features. ECA only captures the local channel interdependencies without relying on global statistics to reduce computational requirements. The advantage of RFA is that it provides effective attention weight to realize convolutional kernel parameter sharing. BRA is a dynamic, query-aware sparse attention mechanism that enables a small subset of the most relevant key/value tokens for each query in a content-aware manner.

We improve the proposed GFPN-based feature fusion structure by adopting the BRA attention module to achieve effective multilevel feature fusion while avoiding redundant information across feature maps. The dynamic sparse attention can reduce redundant feature information and improve the model's detection accuracy, by applying the weight distribution of each channel and spatial position when integrating feature maps of different scales. We place the BRA modules behind the Conv or Upsample module in the feature fusion process to make the model only focus on specific areas after feature extraction. To further avoid information loss, skip connections in CSP modules enable the knowledge of the underlying feature maps to be reused in subsequent layers. BRA aims to eliminate a majority of non-pertinent key-value pairs input at a broader regional level, leaving only a select few relevant areas. Taking a feature map as input, the BRA first segments it into various areas and derives the query, key, and value through a linear transformation. The region-level relationship of queries and keys is entered into an adjacency matrix to construct a directed graph and pinpoint the association of specific key-value pairs. This essentially identifies which areas should be involved with each designated region. Lastly, multi-head self-attention is executed between individual tokens by utilizing the region-to-region routing index matrix. Through the bi-level routing optimization for multi-head self-attention, more attention is paid to the brain tumor part of the feature map, thereby improving the model's ability to detect brain tumors. The BRA attention module can be integrated into feature fusion to allow the model to focus more on relevant regions of the MRI scans where tumors are likely to be present. This selective focus helps in capturing intricate details and subtle features associated with tumors.

This proposed method only uses the attention module BRA of BiFormer, which is different from the existing works [23,35] that add BiFormer into YOLOv8.

### 2.3   An Enhanced Detecting Head

The original YOLOv8 has three detecting heads with respective dimensions in height and width of 20 × 20, 40 × 40, and 80 × 80. In contrast, these heads still cannot meet the detection needs of brain tumor detection scenarios, which leads to the unsatisfactory detection accuracy of the model for larger objects than the original scales.

We introduce an additional 160 ×160 detecting head in the head part aligned with the new structure of feature fusion networks in the neck part to improve the detection capacity for objects in various scales. The new scale-detecting head is added as the fourth detecting head next to the original 80 × 80 detection scales of YOLOv8. It fuses the shallow information of the first C2f (shortcut) module from the input images, incorporating additional feature fusion networks. The one more prediction head we add enhances the model to detect objects in richer scales.

The additional detecting head can classify and localize different brain tumor sizes within the model. In the case of brain tumor detection, this can be beneficial when there are different sizes of tumors with distinct characteristics. Extensive layers of detection heads can facilitate progressive detection, where the model first identifies potential regions of interest and then refines its predictions through subsequent processing stages. This step-wise approach can lead to more accurate and robust tumor detection.

## 3   Experiments and Results

### 3.1   Dataset Details

We evaluated the performance of the proposed BGF-YOLOv8 on the public brain tumor image dataset Br35H [2] which contains 801 MRI images with annotated brain tumors. The dataset was divided into the train set of 500 images, the validation set of 201 images, and the test set of 100 images. All the results are tested on the test set.

### 3.2   Implementation Details

The BGF-YOLO was trained and tested with Intel® Xeon® Platinum 8255C CPU @ 2.50GHz and NVIDIA® GeForce GTX® 1060 6GB GPU. We implemented the proposed methods based on YOLOv8 extra large version (YOLOv8x). The hyperparameters used in the training of BGF-YOLO and other comparison methods are the same as YOLOv8x. The training parameter batch size is set to 5, and the epoch is 120 at the training stage. The optimizer uses the stochastic gradient descent with the initial and final learning rate of 0.01 and momentum of 0.937.

### 3.3 Results

For a fair comparison, we choose the version with the best performance of the competing models and use the same evaluation metrics as those used to evaluate them. As shown in Table 1, BGF-YOLO achieved 1.2%, 4.7% and 0.7% absolute increase in precision, mean average precision $mAP_{50}$ and $mAP_{50:95}$ respectively, compared to YOLOv8x. It also outperforms YOLOv9-E [24], YOLOv10-X [25], RCS-YOLO, and DAMO-YOLO-L$^*$. BGF-YOLO surpasses not only the baseline YOLOv8 model but also a GFPN-neck detector DAMO-YOLO and a high-accuracy and fast detector RCS-YOLO. As shown in supplementary material Sect. 1, in qualitative comparison, brain tumor tumors are more accurately detected by BGF-YOLO than YOLOv8.

**Table 1.** Performance comparison of YOLOv8x, YOLOv9-E, YOLOv10-X, DAMO-YOLO-L$^*$, RCS-YOLO, and our proposed BGF-YOLO. $^*$ in DAMO-YOLO-L$^*$ indicates distillation was employed. The original code of all DAMO-YOLO versions only prints average precision and average recall. The best results are shown in bold.

| Model | Precision | Recall | $mAP_{50}$ | $mAP_{50:95}$ |
|---|---|---|---|---|
| YOLOv8x [8] | 0.907 | 0.881 | 0.927 | 0.646 |
| YOLOv9-E [24] | 0.927 | 0.869 | 0.919 | 0.630 |
| YOLOv10-X [25] | 0.916 | 0.808 | 0.880 | 0.603 |
| RCS-YOLO [9] | 0.908 | 0.885 | 0.878 | 0.580 |
| DAMO-YOLO-L$^*$ [28] | – | – | 0.900 | 0.610 |
| **BGF-YOLO (Ours)** | **0.919** | **0.926** | **0.974** | **0.653** |

### 3.4 Ablation Study

We conducted a series of ablation studies to assess the advantages of incorporating each method in the proposed BGF-YOLO model, and to investigate the effect of using different techniques in each method on the detection performance through the following extensive experiments. See supplementary material Sect. 2: Tables of Ablation Study.

**Ablation Study on Overall Architecture.** We evaluated four incomplete BGF-YOLO models by removing each method respectively. Supplementary material Table 1 shows that BRA, GFPN, the fourth head, and GIoU all contribute to the accuracy improvement of BGF-YOLO. The w/o GFPN means using the original neck structure FPN-PANet of YOLOv8. Adding the fourth detection head gives the most impact on the overall accuracy improvement especially for $mAP_{50}$, followed by GFPN and BRA.

**Effect of Different Multiscale Feature Fusion Structures.** We compared the proposed BGF-YOLO with BBF-YOLO and BAF-YOLO, which respectively replace GFPN with BiFPN and AFPN in the neck part of BGF-YOLO for feature fusion. As shown in Supplementary material Table 2, the precision, $mAP_{50}$, and $mAP_{50:95}$, expect recall of BGF-YOLO with GFPN structure are much higher than the model with BiFPN and AFPN structures.

**Effect of Different Attention Mechanisms.** We investigated different attention mechanisms with the proposed BGF-YOLO model. The first letters of the model names listed in supplementary material Table 3 represent the attention mechanisms used, which means S, E, C, A, R, and B denote SE, ECA, CBAM, CA, RFA, and BRA. BRA gives the largest performance improvement among the different attention mechanisms compared to the other five alternative ones. Meanwhile, CBAM (i.e., CGF-YOLO) ranks second behind BRA (i.e., BGF-YOLO) in terms of $mAP_{50}$ and has higher values in precision than BRA. Although the $mAP_{50:95}$ of ECA (i.e., EGF-YOLO) and CA (i.e., AGF-YOLO) are higher than that of BRA, the $mAP_{50}$ of ECA and CA are much lower than that of BRA.

**Effect of Different Regression Losses.** We performed an ablation study on the influence of regression losses, including Generalized IoU (GIoU) [18] where the distance between two axis-aligned rectangles is calculated, Distance-IoU (DIoU) [36] optimal objectives of which are less than CIoU, Efficient IoU (EIoU) [33] which explicitly measures the discrepancies of three geometric factors, Scylla-IoU (SIoU) [1] where penalty metrics are redefined, WIoU v3 which is a two-layer attention-based with a dynamic nonmonotonic focusing mechanism regression loss function. These loss functions are represented by C, E, S, and W as the fourth letters of model names in supplementary material Table 4. Compared to other regression losses, The original regression loss CIoU in YOLOv8 has better robustness of the bounding box for object detection. The $mAP_{50}$ of DIoU (i.e., BGFD-YOLO) is close to that of CIoU (i.e., BGF-YOLO), which indicates DIoU is a competitor to CIoU. In terms of $mAP_{50:95}$, those of GIoU (i.e., BGFG-YOLO) and EIoU (i.e., BGFE-YOLO) are higher than that of CIoU. Which regression loss is a better choice depends on the criterion of the specific scenario. In this case, we choose $mAP_{50}$ as the main metric for brain tumor detection, and therefore, CIoU is selected as regression loss in the proposed BGF-YOLO.

## 4  Conclusion

We developed a novel BGF-YOLO model building on YOLOv8 for accurate detection of brain tumors from MRI. We show that the object detection capability of YOLOv8 is substantially enhanced by the optimization of the GFPN feature fusion structure, BRA attention mechanism, and adding a detecting head in the BGF-YOLO model. These modifications enable weighted feature fusion at different levels and at richer scales and produce high-quality anchor boxes with dynamic focusing mechanisms. Besides, the proposed modules in BGF-YOLO are better than the other alternative techniques, as shown in a series of experimental evaluations on different feature fusion structures, attention mechanisms, and regression losses. Our proposed BGF-YOLO becomes the current state-of-the-art model on the brain tumor detector dataset Br35H.

**Acknowledgements.** This work was supported by the Monash University Malaysia and the Ministry of Higher Education, Malaysia under Fundamental Research Grant Scheme FRGS/1/2023/ICT02/MUSM/02/1.

**Disclosure of Interests.** The author has no conflict of interest related to the article.